\documentclass[11pt,a4paper]{article}
\usepackage[hyperref]{emnlp2020}
\usepackage{times}
\usepackage{latexsym}
\usepackage{times}
\usepackage{latexsym}
\usepackage{hyperref}
\usepackage{url}
\usepackage{times}
\usepackage{latexsym}
\usepackage{times}
\usepackage{epsfig}
\usepackage{graphicx}
\usepackage{amsmath}
\usepackage{amssymb}
\usepackage{color}
\usepackage{amsmath}
\usepackage{amssymb}
\usepackage{fancyhdr}
\usepackage{listings}
\usepackage{multirow}
\usepackage{graphicx}
\usepackage{verbatim}
\usepackage{bbm}
\usepackage{color}
\usepackage{mathtools}
\usepackage{pifont}
\usepackage{makecell}
\usepackage{multirow}
\usepackage{booktabs}
\usepackage{float}
\usepackage{amsmath,amsfonts,amsthm,bm}
\usepackage{enumitem}
\usepackage{url}
\usepackage{pifont}
\usepackage{xcolor}
\usepackage{sidecap}
\usepackage{booktabs,siunitx} 
\usepackage{tikz}
\usepackage{array}
\usepackage{algorithmic}
\usepackage{algorithm}
\usepackage{subfig}

\newcommand{\Sref}[1]{\S\ref{#1}}

\usepackage{microtype}

\aclfinalcopy 

\title{On Negative Interference in Multilingual Models:\\ Findings and A Meta-Learning Treatment}

\author{Zirui Wang \qquad Zachary C. Lipton \qquad  Yulia Tsvetkov \\
 Carnegie Mellon University, Pittsburgh, USA \\
 {\texttt{\{ziruiw, zlipton, ytsvetko\}@cs.cmu.edu}} }

\date{}

\begin{document}
\maketitle
\begin{abstract}
Modern multilingual models are trained on concatenated text 
from multiple languages in hopes of conferring benefits to each (positive transfer),
with the most pronounced benefits accruing to low-resource languages.
However, recent work has shown that this approach can degrade 
performance on high-resource languages, 
a phenomenon known as \emph{negative interference}.
In this paper, we present the first systematic study of negative interference.
We show that, contrary to previous belief, 
negative interference 
also impacts low-resource languages.
While parameters are maximally shared to learn language-universal structures, 
we demonstrate that language-specific parameters do exist in multilingual models and they are a potential cause of negative interference.
Motivated by these observations, 
we also present a meta-learning algorithm that obtains 
better cross-lingual transferability 
and alleviates negative interference, 
by adding language-specific layers as meta-parameters 
and training them in a manner that explicitly improves 
shared layers' generalization on all languages.
Overall, our results show that negative interference 
is more common than previously known, 
suggesting new directions for improving multilingual representations.\footnote{Source code is available at \url{https://github.com/iedwardwangi/MetaAdapter}.}
%
\end{abstract}

\section{Introduction}

Advances in pretraining language models \cite{devlin2018bert,liu2019roberta,yang2019xlnet} 
as general-purpose representations have pushed 
the state of the art on a variety of natural language tasks.
However, not all languages enjoy large public datasets
for pretraining and/or downstream tasks.
Multilingual language models such as mBERT \cite{devlin2018bert} and XLM \cite{lample2019cross} 
have been proven effective for cross-lingual transfer learning 
by pretraining a single shared Transformer model 
\cite{vaswani2017attention} jointly on multiple languages.
The goals of multilingual modeling are not limited 
to improving language modeling in low-resource languages \cite{lample2019cross}, 
but also include zero-shot cross-lingual transfer on downstream tasks---it 
has been shown that multilingual models can generalize to target languages 
even when labeled training data is only available in the source language (typically English) 
on a wide range of tasks \cite{pires2019multilingual,wu2019beto,hu2020xtreme}. 

However, multilingual models are not equally beneficial for all languages.
\citet{conneau2019unsupervised} demonstrated that including more languages in a single model 
can improve performance for low-resource languages but hurt performance for high-resource languages.
Similarly, recent work \cite{johnson2017google,tan2019multilingual,DBLP:journals/corr/abs-1903-00089,arivazhagan2019massively} 
in multilingual neural machine translation (NMT) also observed 
performance degradation on high-resource language pairs.
In multi-task learning \cite{ruder2017overview}, this phenomenon is known as
\emph{negative interference} or \emph{negative transfer} \cite{wang2019characterizing}, 
where training multiple tasks jointly hinders the performance on individual tasks.

Despite these empirical observations, little prior work analyzed or showed 
how to mitigate negative interference in multilingual language models.
Particularly, it is natural to ask:
(1) Can negative interference occur for low-resource languages also?
(2) What factors play an important role in causing it?
(3) Can we mitigate negative interference to improve the model's cross-lingual transferability?

In this paper, we take a step towards addressing these questions.
We pretrain a set of monolingual and bilingual models and evaluate them 
on a range of downstream tasks to analyze negative interference.
We seek to individually characterize the underlying factors of negative interference 
through a set of ablation studies and glean insights on its causes.
Specifically, we examine if training corpus size and language similarity affect negative interference, 
and also measure gradient and parameter similarities between languages.

Our results show that negative interference can occur in both high-resource and low-resource languages.
In particular, we observe that neither subsampling the training corpus 
nor adding typologically similar languages
substantially impacts negative interference.
On the other hand, we show that gradient conflicts 
and language-specific parameters do exist in multilingual models, 
suggesting that languages are fighting for model capacity,
which potentially causes negative interference.
We further test whether explicitly assigning language-specific modules 
to each language can alleviate negative interference,
and find that the resulting model performs better 
within each individual language but worse on zero-shot cross-lingual tasks.

Motivated by these observations, we further propose 
to meta-learn these language-specific parameters 
to explicitly improve generalization of shared parameters on all languages.
Empirically, our method improves not only within-language performance on monolingual tasks 
but also cross-lingual transferability on zero-shot transfer benchmarks.
To the best of our knowledge, this is the first work to systematically study 
and remedy negative interference in multilingual language models.

\section{Motivation}

\begin{figure}
    \centering
    \includegraphics[width=0.7\columnwidth]{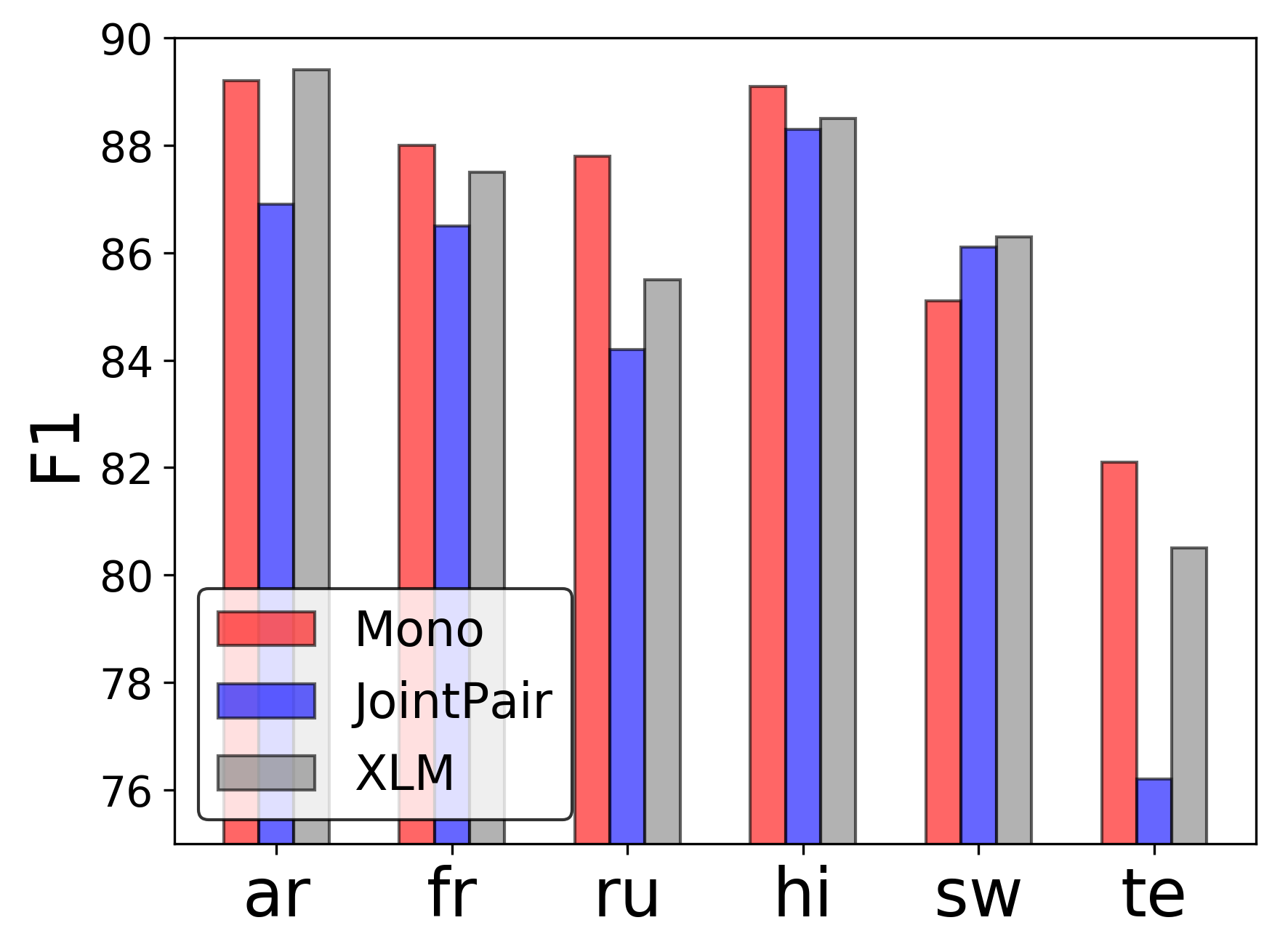}
    \caption{Comparing monolingual vs multilingual models on NER. 
    Lower performance of multilingual models is likely an indicator of negative interference. }
    \label{fig:summary}
\vskip -0.1in
\end{figure}

Multilingual transfer learning aims at utilizing knowledge transfer across languages
to boost performance on low-resource languages.
State-of-the-art multilingual language models are trained on multiple languages jointly 
to enable cross-lingual transfer through parameter sharing.
However, languages are heterogeneous, 
with different vocabularies, morphosyntactic rules, 
and different pragmatics across cultures.
It is therefore natural to ask, 
\emph{is knowledge transfer beneficial for all languages in a multilingual model?}
To analyze the effect of knowledge transfer 
from other languages on a specific language $\text{\it lg}$, 
we can compare multilingual models with 
the monolingual model trained on $\text{\it lg}$. 
For example, in Figure \ref{fig:summary}, 
we compare the performance on a named entity recognition (NER) task 
of monolingually-trained models 
vs.~bilingual models 
(trained on $\text{\it lg}$ and English) 
vs.~state-of-the-art XLM \cite{conneau2019unsupervised}.
We can see that monolingual models outperform multilingual models 
on four out of six languages (See \Sref{sec:eval} for details).
This shows that language conflicts may induce 
negative impacts on certain languages, 
which we refer to as \emph{negative interference}.
Here, we investigate the causes of negative interference~(\Sref{sec:eval}) 
and methods to overcome it~(\Sref{sec:method}).

\section{Investigating the Sources of Negative Interference in Multilingual Models}

\subsection{Methodology}
\label{sec:setup}

To study negative interference, we compare multilingual models with monolingual baselines.
Without loss of generality, we focus on analyzing bilingual models to minimize confounding factors.
For two languages $\text{\it lg}_1$ and $\text{\it lg}_2$, 
we pretrain a single bilingual model and two monolingual models.
We then assess their performance on downstream tasks using two different settings.
To examine negative interference, we evaluate both monolingual and multilingual models 
using the \textbf{within-language monolingual} setting, 
such that the pretrained model is finetuned and tested on the same language.
For instance, if the monolingual model of $\text{\it lg}_1$
outperforms the bilingual model on $\text{\it lg}_1$,
we know that $\text{\it lg}_2$ induces negative impact 
on $\text{\it lg}_1$ in the bilingual model.  
Besides, since multilingual models are trained to enable cross-lingual transfer, 
we also report their performance on the \textbf{zero-shot cross-lingual transfer} setting, 
where the model is only finetuned on the source language, 
say $\text{\it lg}_1$, and tested on the target language $\text{\it lg}_2$.

We hypothesize that the following factors play important roles 
in causing negative interference and study each individually:

\vspace{5px}
\noindent \textbf{Training Corpus Size \quad}
While prior work mostly report negative interference for high-resource languages \cite{conneau2019unsupervised,arivazhagan2019massively}, 
we hypothesize that it can also occur for languages with less resources.
We study the impact of training data size per language on negative interference.
We subsample a high-resource language, say $\text{\it lg}_1$, to create a ``low-resource version''.
We then retrain the monolingual and bilingual models and compare with results of their high-source counterparts.
Particularly, we test if reducing $\text{\it lg}_1$'s training size
also reduces negative interference on $\text{\it lg}_2$. 

\noindent \textbf{Language Similarity \quad}
Language similarity has been shown important for effective transfer in multilingual models.
\citet{wu2019emerging} shows that bilingual models 
trained on more similar language pairs 
result in better zero-shot transfer performance.
We thus expect it to play a critical role in negative interference as well.
For a specific language $\text{\it lg}_1$, we pair it with languages that are closely and distantly related.
We then compare these bilingual models' performance on $\text{\it lg}_1$ 
to investigate if more similar languages cause less severe interference.
In addition, we further add a third language $\text{\it lg}_3$ 
that is similar to $\text{\it lg}_1$ and train a trilingual model 
on $\text{\it lg}_1$-$\text{\it lg}_2$-$\text{\it lg}_3$.
We compare the trilingual model with the bilingual model 
to examine if adding $\text{\it lg}_3$ can mitigate negative interference on $\text{\it lg}_1$.

\noindent \textbf{Gradient Conflict \quad}
Recent work \cite{yu2020gradient} shows that gradient conflict between dissimilar tasks, 
defined as a negative cosine similarity between gradients,
is predictive of negative interference in multi-task learning.
Therefore, we study whether gradient conflicts exist between languages in multilingual models.
In particular, we sample one batch for each language in the model 
and compute the corresponding gradients' cosine similarity 
for every 10 steps during pretraining.

\begin{table}[t!]
\begin{center}
\resizebox{0.9\columnwidth}{!}{%
\begin{tabular}{l c c c c c c c }
\toprule
& en & ar & fr & ru & hi & sw & te \\
\toprule
corpus size & 44.6 & 8.7 & 16.2 & 13.1 & 0.5 & 0.2 & 0.3\\
NER & \checkmark & \checkmark & \checkmark & \checkmark & \checkmark & \checkmark & \checkmark \\
POS & \checkmark & \checkmark & \checkmark & \checkmark & \checkmark & & \checkmark\\
QA & \checkmark & \checkmark & & \checkmark & & \checkmark & \checkmark\\
XNLI & \checkmark & \checkmark & \checkmark & \checkmark & \checkmark & \checkmark & \\
\bottomrule
\end{tabular}%
}
\end{center}
\caption[caption]{Language training corpra statitstics and downstream tasks availability. Corpus size measured in millions of sentences.}
\vskip -0.1in
\label{tab:stats}
\end{table}

\begin{table*}[t!]
\begin{center}
\begin{small}
\begin{tabular}{l | c c c c c c c | c c c c c c }
\toprule
    \multirow{2}{*}{Model} & \multicolumn{7}{c|}{NER (F1)} & \multicolumn{6}{c}{POS (F1)} \\
    & ar & fr & ru & hi & sw & te & avg & ar & fr & ru & hi & te & avg \\
\toprule
\multicolumn{14}{c}{Within-language Monolingual}\\
\toprule
Mono & 89.2 & 88.0 & 87.8 & 89.1 & 85.1 & 82.1 & 86.9 & 92.7 & 76.2 & 96.7 & 97.0 & 94.5 & 91.4 \\
JointPair & 86.9 & 86.5 & 84.2 & 88.3 & 86.1 & 76.2 & 84.7 & 89.2 & 75.8 & 93.2 & 95.2 & 88.7 & 88.4\\
\enspace + ffn & 88.2 & 88.4 & 86.6 & 88.9 & 85.4 & 81.2 & 86.5 & 92.4 & 76.1 & 95.6 & 96.1 & 92.4 & 90.5\\
\enspace + attn & 87.3 & 86.8 & 84.1 & 88.5 & 84.9 & 77.4 & 84.8 & 91.8 & 75.4 & 94.4 & 95.3 & 90.9 & 89.6\\
\enspace + adpt & 87.8 & 86.8 & 84.5 & 87.7 & 86.3 & 77.0 & 85.0 & 91.7 & 75.6 & 94.0 & 95.2 & 91.5 & 89.6 \\
\enspace + share adpt & 86.8 & 86.7 & 84.3 & 88.6 & 86.1 & 76.0 & 84.8 & 89.3 & 76.4 & 93.5 & 95.2 & 88.2 & 88.5\\
\enspace + meta adpt & 88.9 & 88.3 & 85.1 & 88.4 & 86.5 & 79.5 & 86.1 & 92.4 & 75.9 & 95.1 & 95.8 & 92.2 & 90.3 \\
\toprule
XLM & 89.4 & 87.5 & 85.5 & 88.5 & 86.3 & 80.5 & 86.3 & 94.5 & 72.9 & 96.6 & 97.1 & 92.2 & 90.7\\
\toprule
\multicolumn{14}{c}{Zero-shot Cross-lingual}\\
\toprule
JointPair & 38.1 & 77.5 & 57.5 & 61.4 & 64.8 & 45.2 & 57.4 & 58.5 & 44.2 & 80.1 & 58.9 & 72.8 & 62.9 \\
\enspace + ffn & 8.9 & 35.2 & 5.8 & 10.5 & 9.7 & 12.5 & 13.8 & 5.4 & 8.1 & 4.5 & 3.3 & 7.7 & 5.8 \\
\enspace + attn & 15.4 & 39.4 & 10.2 & 9.9 & 13.4 & 11.6 & 16.7 & 6.2 & 4.5 & 7.5 & 4.8 & 6.9 & 6.0 \\
\enspace + adpt & 37.2 & 75.5 & 59.2 & 61.0 & 64.4 & 44.7 & 57.0 & 57.0 & 43.5 & 81.6 & 58.2 & 73.5 & 62.8 \\
\enspace + share adpt & 38.5 & 77.8 & 58.4 & 62.0 & 65.4 & 44.5 & 57.8 & 58.7 & 43.8 & 82.5 & 59.7 & 71.8 & 63.3 \\
\enspace + meta adpt & 44.4 & 78.5 & 62.4 & 66.0 & 67.3 & 50.1 & 61.5 & 63.5 & 44.6 & 84.9 & 62.7 & 78.5 & 66.8 \\
\toprule
XLM & 44.8 & 78.3 & 63.6 & 65.8 & 68.4 & 49.3 & 61.7 & 62.8 & 42.4 & 86.3 & 65.7 & 76.9 & 66.8 \\
\bottomrule
\end{tabular}
\end{small}
\end{center}
\caption[caption]{NER and POS results. We observe negative interference when monolingual models outperform multilingual models. Besides, adding language-specific layers (e.g. ffn) mitigates interference but sacrifices transferability.}
\vskip -0.1in
\label{tab:ner_pos}
\end{table*}

\noindent \textbf{Parameter Sharing \quad}
State-of-the-art multilingual models aim to share as many parameters
as possible
in the hope of learning a language-universal model for all languages \cite{wu2019emerging}.
While prior studies measure the latent embedding similarity between languages, 
we instead examine model parameters directly.  
The idea is to test whether model parameters are
\textbf{language-universal} or \textbf{language-specific}.
To achieve this, we prune multilingual models for each language using relaxed $L_0$ norm regularization \cite{louizos2017learning}, and compare parameter similarities between languages.
Formally, for a model $f(\cdot; \boldsymbol{\theta})$ 
parameterized by $\boldsymbol{\theta}=\{\theta_i\}_{i=1}^{n}$
where each $\theta_i$ represents an individual parameter or a group of parameters, 
the method introduces a set of binary masks $\mathbf{z}$,
drawn from some distribution $q(\mathbf{z}|\boldsymbol{\pi})$ 
parametrized by $\boldsymbol{\pi}$, and learns a sparse model
$f(\cdot; \boldsymbol{\theta} \odot \mathbf{z})$ by optimizing:  
\begin{equation}
\begin{aligned}
\min_{\boldsymbol{\pi}} \quad & \mathbb{E}_{q(\mathbf{z}|\boldsymbol{\pi})} \left[\frac{1}{N} \sum_{i=1}^{N} \mathcal{L}(f(x_i; \tilde{\boldsymbol{\theta}}), y_i) + \lambda \| \tilde{\boldsymbol{\theta}} \|_0\right]\\
\text{s.t.} \quad & \tilde{\boldsymbol{\theta}} = \boldsymbol{\theta} \odot \mathbf{z},
\end{aligned}
\end{equation}
where $\odot$ is the Hadamard (elementwise) product, 
$\mathcal{L}(\cdot)$ is some task loss and $\lambda$ is a hyper-parameter.
We follow the work of \cite{louizos2017learning} and use 
the Hard Concrete distribution for the binary mask $\mathbf{z}$,
such that the above objective is fully differentiable.
Then, for each bilingual model, we freeze its pretrained parameter weights 
and learn binary masks $\mathbf{z}$ for \textit{each} language independently.
As a result, we obtain two independent sets of mask parameters $\boldsymbol{\pi}$ 
which can be used to determine parameter importance.
Intuitively, for each parameter group, it is language-universal 
if both languages consider it important (positive $\boldsymbol{\pi}$ values).
On the other hand, if one language assigns positive value while the other assigns negative,
it shows that the parameter group is language-specific. 
We compare them across languages and layers to analyze parameter similarity in multilingual models.

\subsection{Experimental Setup}
We focus on standard multilingual masked language modeling (MLM) used in mBERT and XLM.
We first pretrain models and then evaluate their performance on four NLP benchmarks.

For pretraining, we mainly follow the setup and implementation of XLM \cite{lample2019cross}.
We focus on monolingual and bilingual models for a more controllable comparison, 
which we refer to as \textbf{Mono} and \textbf{JointPair} respectively.
In particular, we always include English (En) in bilingual models to compare on zero-shot transfer settings with prior work.
Besides, we consider three high-resource languages 
\{Arabic (Ar), French (Fr), Russian (Ru)\} and 
three low-resource languages \{Hindi (Hi), Swahili (Sw), Telugu (Te)\} (see Table \ref{tab:stats} for their statistics).
We choose these six languages based their data availability in downstream tasks.
We use Wikipedia as training data with statistics shown in Table \ref{tab:stats}.
For each model, we use BPE \cite{sennrich2016neural} 
to learn 32k subword vocabulary shared between languages.
For multilingual models, we sample language proportionally to $P_i = (\frac{L_i}{\sum_j L_j})^{\frac{1}{T}}$, where $L_i$ is the size of the training corpus for $i$-th language pair and T is the temperature.
Each model is a standard Transformer \cite{vaswani2017attention} 
with 8 layers, 12 heads, 512 embedding size and 2048 hidden dimension for the feedforward layer.
Notice that we specifically consider a smaller model capacity 
to be comparable with existing models with larger capacity 
but also include much more (over 100) languages.
We use the Adam optimizer \cite{kingma2014adam} and exploit 
the same learning rate schedule as \citet{lample2019cross}.
We train each model with 4 NVIDIA V100 GPUs with 32GB of memory.
Using mixed precision, we fit a batch of 128 for each GPU and the total batch size is 512.
Each epoch contains 10k steps and we train for 50 epochs.

For evaluation, we consider four downstream tasks: 
named entity recognition (NER), part-of-speech tagging (POS), 
question answering (QA), and natural language inference (NLI).
(See Appendix \ref{sec:finetune_details} for finetuning details.)

\noindent \textbf{NER \quad} We use the WikiAnn \cite{pan2017cross} dataset, 
which is a sequence labelling task built automatically from Wikipedia. 
A linear layer with softmax classifier is added on top of pretrained models 
to predict the label for each word based on its first subword.
We report the F1 score.

\noindent \textbf{POS \quad} Similar to NER, POS is also a sequence labelling task but with a focus on synthetic knowledge.
In particular, we use the Universal Dependencies treebanks \cite{nivre2018universal}.
Task-specific layers are the same and we report F1, as in NER.

\noindent \textbf{QA \quad} We choose to use the TyDiQA-GoldP dataset \cite{clark2020tydi} 
that covers typologically diverse languages. 
Similar to popular QA dataset such as SQuAD \cite{rajpurkar2018know},
this is a span prediction task where task-specific linear classifiers are used 
to predict start/end positions of the answer.
Standard metrics of F1 and Exact Match (EM) are reported.

\noindent \textbf{NLI \quad} XNLI \cite{conneau2018xnli} is probably the most popular cross-lingual benchmark.
Notice that the original dataset only contains training data for English.
Consequently, we only evaluate this task on the zero-shot transfer setting 
while we consider both settings for the rest of other tasks.

\begin{table}[t!]
\begin{center}
\begin{small}
\resizebox{0.9\columnwidth}{!}{%
\begin{tabular}{l | c c c c c }
\toprule
    Model & ar  & ru  & sw & te & avg \\
\toprule
\multicolumn{6}{c}{Within-language Monolingual}\\
\toprule
Mono & 74.2 & 63.1 & 52.5 & 58.2 & 62.0 \\
JointPair & 71.3 & 58.2 & 52.8 & 52.2 & 58.6  \\
\enspace + ffn & 73.4 & 61.2 & 51.4 & 57.5 & 60.9 \\
\enspace + attn & 72.8 & 60.8 & 51.2 & 52.8 & 59.4 \\
\enspace + adpt & 71.5 & 59.4 & 52.1 & 55.5 & 59.6 \\
\enspace + share adpt & 71.0 & 58.5 & 52.8 & 53.9 & 59.1 \\ 
\enspace + meta adpt & 73.0 & 61.8 & 54.5 & 56.2 & 61.4 \\
\toprule
XLM & 74.3 & 62.5 & 58.7 & 55.4 & 62.7 \\
\toprule
\multicolumn{6}{c}{Zero-shot Cross-lingual}\\
\toprule
JointPair & 54.1 & 43.2 & 41.5 & 21.5 & 40.1 \\
\enspace + ffn & 2.2 & 0.0 & 4.4 & 0.0 & 1.7 \\
\enspace + attn & 3.7 & 2.1 & 0.7 & 0.0 & 1.6 \\
\enspace + adpt & 53.4 & 44.7 & 41.2 & 20.4 & 39.9 \\
\enspace + share adpt & 54.3 & 44.8 & 42.2 & 22.7 & 41.0 \\
\enspace + meta adpt & 57.5 & 45.8 & 43.0 & 23.1 & 42.4 \\
\toprule
XLM & 59.4 & 47.3 & 42.3 & 16.3 & 41.3 \\
\bottomrule
\end{tabular}%
}
\end{small}
\end{center}
\caption[caption]{TyDiQA-GoldP results (F1). See Appendix \ref{sec:extra_results} for full results.}
\vskip -0.1in
\label{tab:qa}
\end{table}

\subsection{Results and Analysis}
\label{sec:eval}

In Table \ref{tab:ner_pos} and \ref{tab:qa}, 
we report our results on NER, POS and QA together with XLM-100, 
which is trained on 100 languages and contains 827M parameters.
In particular, we observe that monolingual models outperform bilingual models 
for all languages except Swahili on all three tasks. 
In fact, monolingual models even perform better than XLM 
on four out of six languages including \emph{hi} and \emph{te}, 
despite that XLM is much larger in model sizes and trained with much more resources.
This shows that negative interference 
\emph{can} occur on low-resource languages as well.
While the negative impact is expected to be more prominent on high-resource languages,
we demonstrate that it may occur for languages with resources fewer than commonly believed. 
The existence of negative interference confirms 
that state-of-the-art multilingual models cannot generalize equally well on all languages, 
and there is still a gap compared to monolingual models on certain languages.

We next turn to dissect negative interference
by studying the four factors described in Section \ref{sec:setup}.

\begin{figure}%
    \centering
    \subfloat[hi]{\includegraphics[width=0.24\textwidth]{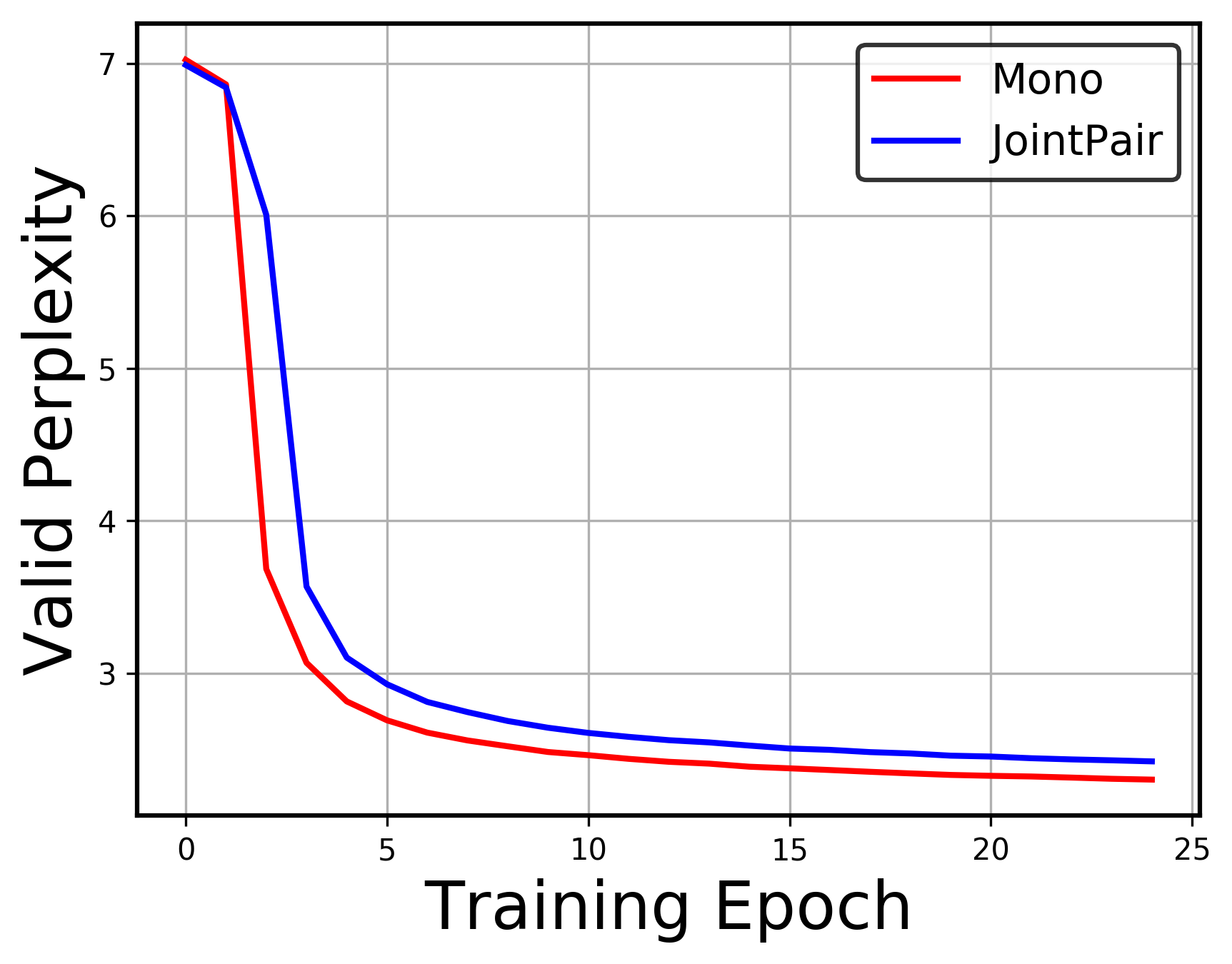} }%
    \subfloat[sw]{\includegraphics[width=0.24\textwidth]{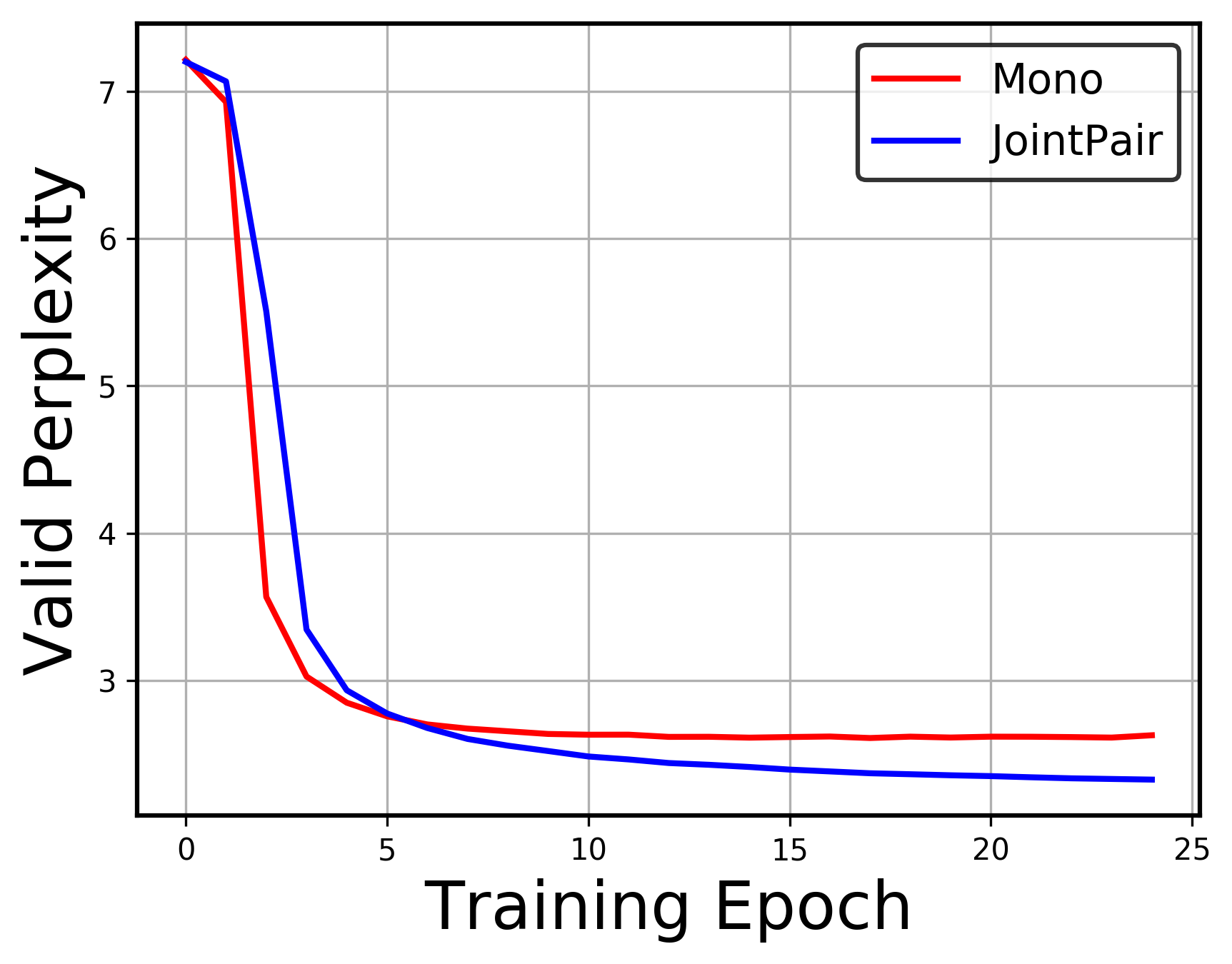} }
    \caption{Validation perplexity during pretraining.}
    \label{fig:ppl}%
\vskip -0.1in
\end{figure}

\vspace{5px}
\noindent \textbf{Training Corpus Size \quad}
By comparing the validation perplexity on Swahili and Telugu in Figure \ref{fig:ppl},
we find that while both monolingual models outperform bilingual models in the first few epochs, 
the Swahili model's perplexity starts to \textit{increase} 
and is eventually surpassed by the bilingual model in later epochs.
This matches the intuition that monolingual models may overfit when training data size is small.
To verify this, we subsample French and Russian to 100k sentences 
to create a ``low-resource version" of them (denoted as $\text{fr}_{l}$/$\text{ru}_{l}$).
As shown in Table \ref{tab:data_size}, while the performance 
for both models drop compared to their ``high-resource" counterparts, 
bilingual models indeed outperform monolingual models for $\text{fr}_{l}$/$\text{ru}_{l}$, 
in contrast for fr/ru.
This suggests that multilingual models can stimulate positive transfer 
for low-resource languages when monolingual models overfit.
On the other hand, when we compare bilingual models on English, 
models trained using different sizes of fr/ru data obtain similar performance, 
indicating that the training size of the source language 
has little impact on negative interference on the target language (English in this case).
While more training data usually implies larger vocabulary 
and more diverse linguistic phenomena, 
negative interference seems to arise from more fundamental conflicts 
contained in even small training corpus.

\begin{table}[t!]
\begin{center}
\begin{small}
\begin{tabular}{l | c c | c c }
\toprule
\multirow{2}{*}{Model} & \multicolumn{2}{c|}{NER (F1)} & \multicolumn{2}{c}{POS (F1)} \\
    & hi & te & hi & te \\
\toprule
\multicolumn{5}{c}{Within-language Monolingual}\\
\toprule
JointPair & 88.3 & 76.2 & 95.2 & 88.7 \\
JointTri & 87.8 & 76.4 & 95.3 & 88.7 \\
\toprule
\multicolumn{5}{c}{Zero-shot Cross-lingual}\\
\toprule
JointPair & 61.4 & 45.2 & 58.9 & 72.8 \\
JointTri & 63.5 & 47.6 & 59.5 & 74.4 \\
\bottomrule
\end{tabular}
\end{small}
\end{center}
\caption[caption]{Comparing trilingual models with bilingual models. This shows the effect of adding a third similar language to bilingual models.}
\vskip -0.1in
\label{tab:tri_models}
\end{table}

\begin{table*}
\begin{center}
\begin{small}
\begin{tabular}{l | c c c c | c c c c | c c }
\toprule
\multirow{2}{*}{Model} & \multicolumn{4}{c|}{NER (F1)} & \multicolumn{4}{c|}{POS (F1)} & \multicolumn{2}{c}{QA (F1/EM)} \\
    & fr & $\text{fr}_{l}$ & ru & $\text{ru}_{l}$ & fr & $\text{fr}_{l}$ & ru & $\text{ru}_{l}$ & ru & $\text{ru}_{l}$ \\
\toprule
\multicolumn{11}{c}{Within-language Performance on fr/ru}\\
\toprule
Mono & 88.0 & 81.7 & 87.8 & 82.4 & 76.2 & 68.5 & 96.7 & 88.7 & 63.1/49.2 & 47.2/29.5\\
JointPair & 86.5 & 83.2 & 84.2 & 82.7 & 75.8 & 71.4 & 93.2 & 89.5 & 58.2/43.1 & 49.5/30.4\\
\toprule
\multicolumn{11}{c}{Within-language Performance on en}\\
\toprule
JointPair & 78.6 & 78.4 & 75.8 & 75.9 & 94.5 & 94.5 & 92.7 & 92.3 & 61.7/49.8 & 62.1/50.2 \\
\bottomrule
\end{tabular}
\end{small}
\end{center}
\caption[caption]{Evaluating effects of training corpus sizes on negative interference.}
\vskip -0.1in
\label{tab:data_size}
\end{table*}

\noindent \textbf{Language Similarity \quad}
As illustrated by Table \ref{tab:data_size}, the in-language performance on English drops 
as the paired language becomes more distantly related (French vs Russian).
This verifies that transferring from more distant languages 
results in more severe negative interference.

It is therefore natural to ask if adding more similar languages
can mitigate negative interference, especially for low-resource languages.
We then train two trilingual models, 
adding Marathi to English-Hindi, 
and Kannada to English-Telugu.
Compared to their bilingual counterparts (Table~\ref{tab:tri_models}), 
trilingual models obtain similar within-language performance, 
which indicates that adding similar languages
\emph{cannot} mitigate negative interference.
However, they do improve zero-shot cross-lingual performance.
One possible explanation is that even similar languages 
can fight for language-specific capacity but they may 
nevertheless benefit the generalization of the shared knowledge.

\begin{figure}[t!]
    \centering
    \includegraphics[width=0.7\columnwidth]{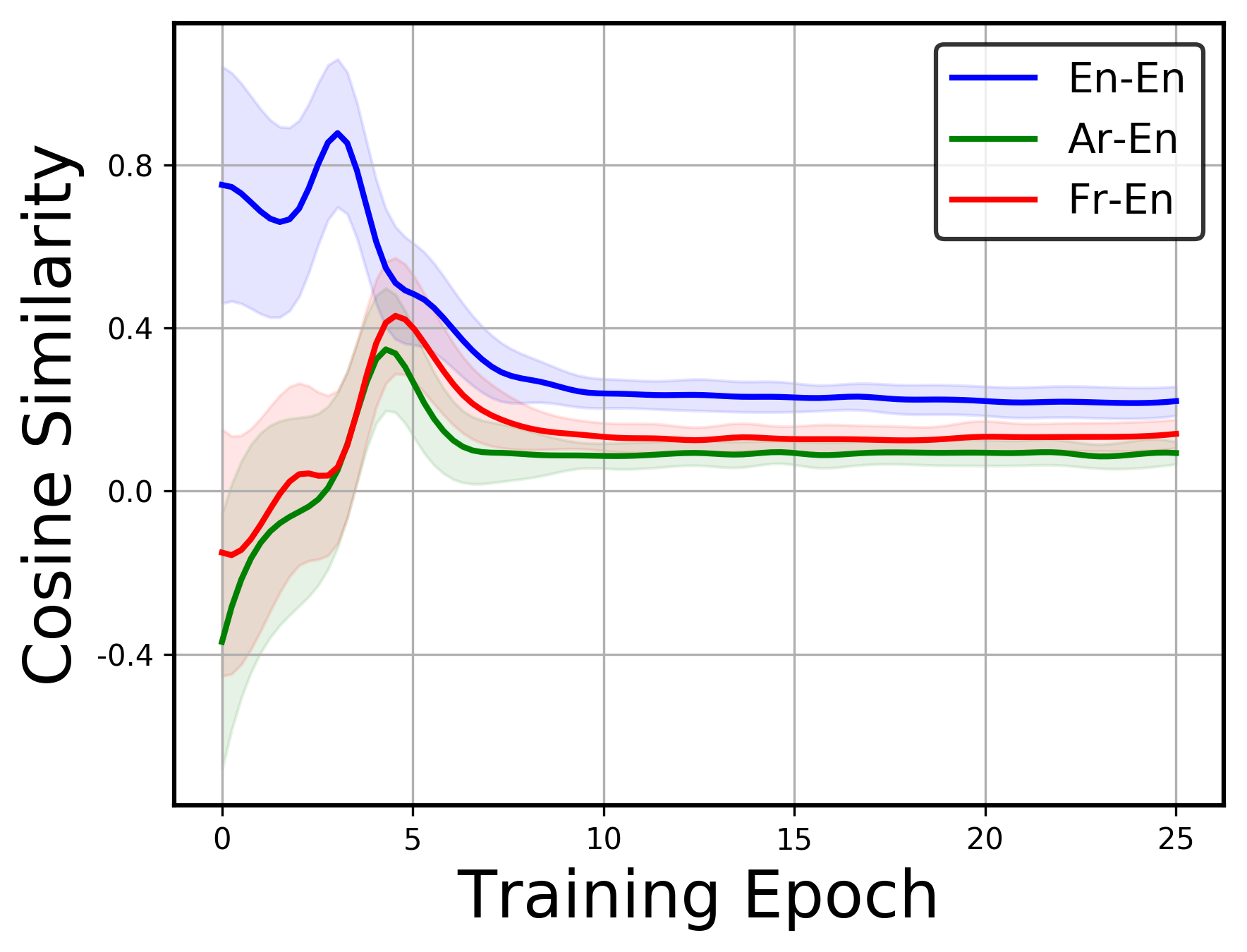}
    \caption{Gradients similarity throughout training. ``En-En'' refers to gradients of two English batches within the Ar-En model, while ``Ar-En'' and ``Fr-En'' refer to gradients of two batches, one from each language, within Ar-En and Fr-En models respectively.}
    \label{fig:grad}
\vskip -0.1in
\end{figure}

\noindent \textbf{Gradient Conflict \quad}
In Figure \ref{fig:grad}, we plot the gradient cosine similarity 
between Arabic-English and French-English in their corresponding bilingual models
over the first 25 epochs.
We also plot the similarity within English, 
measured using two independently sampled batches\footnote{Notice that we use gradient accumulation to sample an effectively larger batch of 4096 sentences to calculate the gradient similarity.}.
Specifically, gradients between two different languages 
are indeed less similar than those within the same language.
The gap is more evident in the early few epochs, 
where we observe negative gradient similarities 
for Ar-En and Fr-En while those for En-En are positive.
In addition, gradients in Ar-En are less similar than those in Fr-En, 
indicating that distant language pair can cause more severe gradient conflicts.
These results confirm that gradient conflict exists in multilingual models 
and is correlated to per language performance, 
suggesting it may introduce optimization challenge 
that results in negative interference. 

\begin{figure*}[h]%
    \centering
    \subfloat[]{\includegraphics[width=0.3\textwidth]{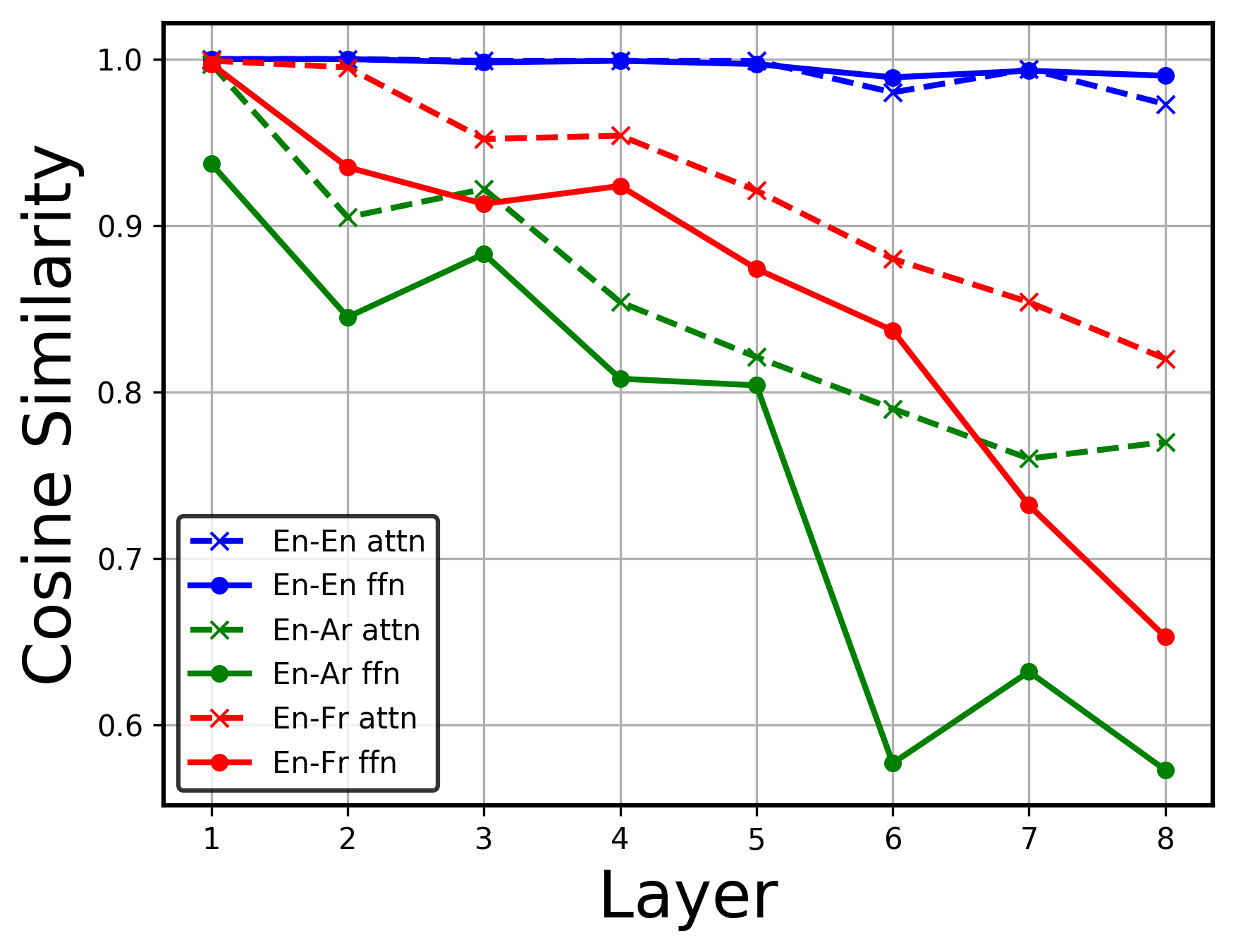} \label{fig:param_cos} }
    \subfloat[]{\includegraphics[width=0.3\textwidth]{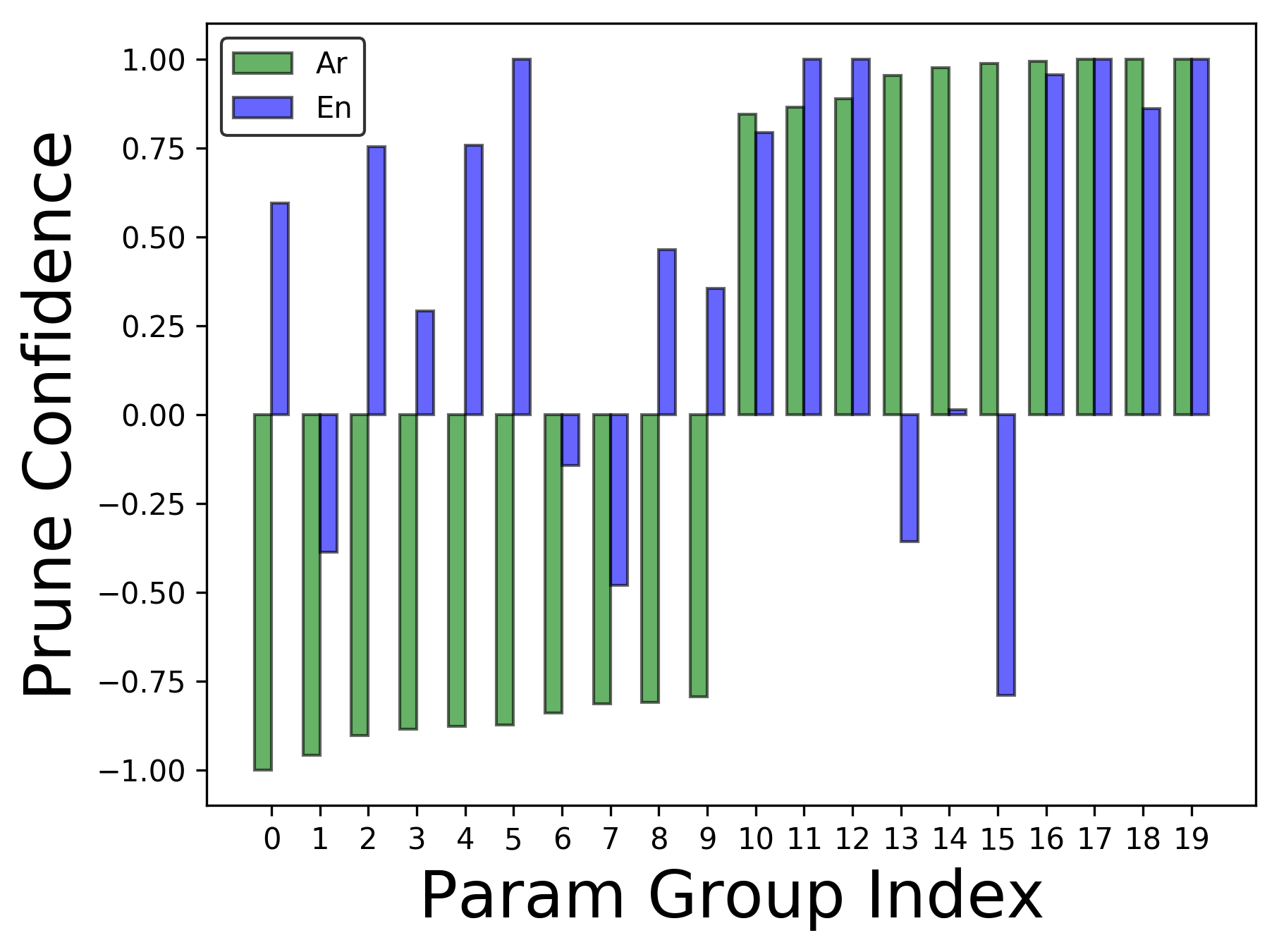} \label{fig:param_visual} }
    \subfloat[]{\includegraphics[width=0.3\textwidth]{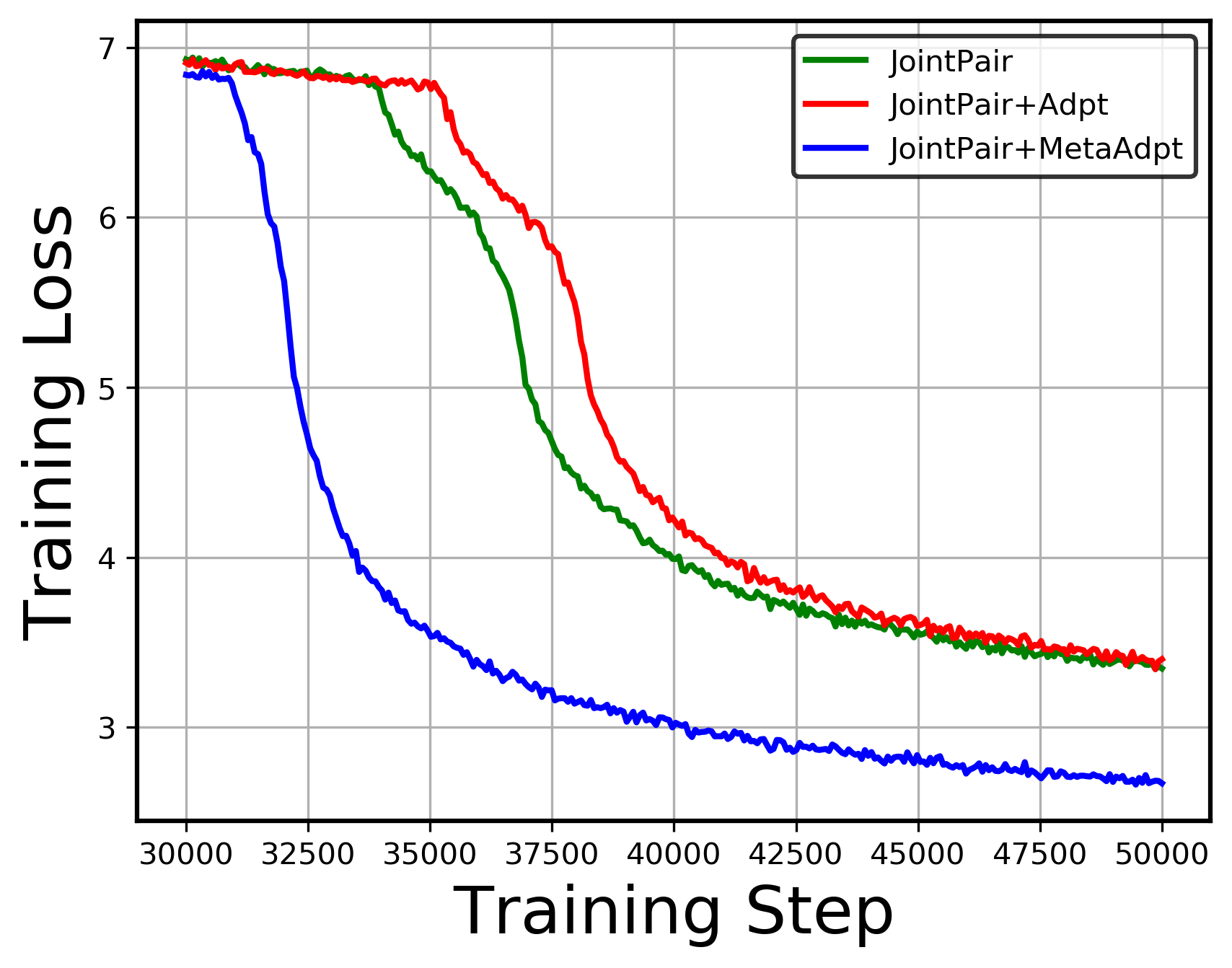} \label{fig:train_loss} }
    \caption{\textbf{Left:} Parameter similarity across layers. \textbf{Middle:} Normalized pruning variables of highest absolute values for Ar in Ar-En model. 10 parameter groups with most negative values are shown on the left and 10 with most positive values are shown on the right. \textbf{Right:} Average MLM training loss after the warm-up stage.}
\vskip -0.1in
\end{figure*}

\noindent \textbf{Parameter Sharing \quad}
The existence of gradient conflicts may imply that languages are fighting for capacity.
Thus, we next study how language-universal these multilingual parameters are.
Figure \ref{fig:param_cos} shows the cosine similarity 
of mask parameters $\boldsymbol{\pi}$ across different layers.
We observe that within-language similarity (En-En) is near perfect,
which validates the pruning method's robustness.
The trend shows that model parameters are better shared 
in the bottom layers than the upper ones.
Besides, it also demonstrates that parameters in multi-head attention layers 
obtain higher similarities than those in feedforward layers, 
suggesting that attention mechanism might be more language-universal.
We additionally inspect $\boldsymbol{\pi}$ parameters with the highest absolute values 
and plot those values for Ar (Figure~\ref{fig:param_visual}), 
together with their En counterparts.
A more negative value indicates that the parameter 
is more likely to be pruned for that language and vice versa.
Interestingly, while many parameters with positive values (on the right) 
are language-universal as both languages assign very positive values, 
parameters with negative values (on the left) 
are mostly language-specific for Ar as En assigns positive values.
We observe similar patterns for other languages as well.
These results demonstrate that language-specific parameters do exist in multilingual models.

Having language-specific capacity in shared parameters is sub-optimal.
It is less transferable and thus can hinder cross-lingual performance.
Moreover, it may also take over capacity budgets for other languages 
and degrade their within-language performance, i.e., causing negative interference.
A natural next question is whether explicitly adding language-specific capacity
into multilingual models can alleviate negative interference.
We thus train variants of bilingual models 
that contain language-specific components for each language.
Particularly, we consider adding language-specific feedforward layers, 
attention layers, and residual adapter layers 
\cite{rebuffi2017learning,houlsby2019parameter}, 
denoted as ffn, attn and adpt respectively.
For each type of component, we create two separate copies in each Transformer layer, 
one designated for each language, while the rest of the network remains unchanged.
As shown in Table \ref{tab:ner_pos} and \ref{tab:qa}, 
adding language-specific capacity does mitigate negative interference 
and improve monolingual performance.
We also find that language-specific feedforward layers 
obtain larger performance gains compared to attention layers, 
consistent with our prior analysis.
However, these gains come at a cost of cross-lingual transferability, 
such that their zero-shot performance drops tremendously.
%
Our results suggest a tension between addressing interference 
versus improving transferability.
In the next section, we investigate how to address negative interference 
in a manner that can improve performance on 
\emph{both} within-language tasks and cross-lingual benchmarks.

\begin{algorithm*}[t!]
\caption{Training XLM with Meta Language-specific Layers}
\label{algorithm}
\begin{algorithmic}[1]
    \STATE {\bfseries Input:} Training data
    \STATE {\bfseries Output:} The converged model \{$\boldsymbol{\theta}^*,\boldsymbol{\phi}^*$\}
    \STATE Initialize model parameters \{$\boldsymbol{\theta}^{(0)},\boldsymbol{\phi}^{(0)}$\}
    \WHILE{not converged}
    \STATE Sample language $i$
    \STATE Update language-specific parameters as: \par
        \hskip\algorithmicindent $\phi_i^{(t+1)} \leftarrow \text{GradientUpdate}(\phi_i^{(t)}, \nabla_{\phi_{i}^{(t)}} \frac{1}{L} \sum_{j=1}^{L} \mathcal{L}_\text{val}^{j}(\theta_i^{(t)} - \beta \nabla_{\boldsymbol{\theta}^{(t)}}\mathcal{L}_\text{train}^{i}(\boldsymbol{\theta}^{(t)}, \phi_{i}^{(t)}), \phi_{j}^{(t)}))$
    \STATE Update shared parameters as: \par
        \hskip\algorithmicindent $\boldsymbol{\theta}^{(t+1)} \leftarrow \text{GradientUpdate}( \boldsymbol{\theta}^{(t)}, \nabla_{\boldsymbol{\theta}^{(t)}} \mathcal{L}_\text{train}(\boldsymbol{\theta}^{(t)}, \boldsymbol{\phi}^{(t+1)}))$
    \ENDWHILE
\end{algorithmic}
\end{algorithm*}

\section{Mitigating Negative Interference via Meta Learning}
\label{sec:method}

\subsection{Proposed Method}

In the previous section, we demonstrated
that while explicitly adding language-specific components 
can alleviate negative interference, 
it can also hinder cross-lingual transferability.
We notice that a critical shortcoming of language-specific capacity
is that they are \textbf{agnostic} of the rest of other languages, 
since by design they are trained on the designated language only. 
They are thus more likely to overfit and can induce 
optimization challenges for shared capacity as well.
Inspired by recent work in meta learning \cite{flennerhag2019meta} 
that utilizes meta parameters to improve gradient geometry of the base network,
we propose a novel meta-learning formulation of multilingual models 
that exploits language-specific parameters to improve generalization of shared parameters.

For a model with some predefined language-specific parameters 
$\boldsymbol{\phi} = \{\phi_i\}_{i=1}^{L}$,
where $\phi_i$ is designated for the i-th language,
and shared parameters $\boldsymbol{\theta}$, 
our solution is to treat $\boldsymbol{\phi}$ as meta parameters 
and $\boldsymbol{\theta}$ as base parameters.
Ideally, we want $\boldsymbol{\phi}$ to store 
non-transferable language-specific knowledge 
to resolve conflicts and improve generalization of $\boldsymbol{\theta}$ 
in all languages (a.k.a. mitigate negative interference and improve cross-lingual transferability).
Therefore, we train $\boldsymbol{\phi}$ based on the following principle:
\textit{if $\boldsymbol{\theta}$ follows the gradients on training data 
for a given  $\boldsymbol{\phi}$, 
the resulting $\boldsymbol{\theta}$ should obtain 
a good validation performance on all languages}.
This implies a bilevel optimization problem \cite{colson2007overview} 
formally written as:
\begin{equation}
\begin{aligned}
\min_{\boldsymbol{\phi}} \quad & \frac{1}{L} \sum_{i=1}^{L} \mathcal{L}_\text{val}^{i}(\boldsymbol{\theta}^*, \phi_i) \\
\text{s.t.} \quad & \boldsymbol{\theta}^* = \arg\min_{\boldsymbol{\theta}} \frac{1}{L} \sum_{i=1}^{L} \mathcal{L}_\text{train}^{i}(\boldsymbol{\theta}, \phi_i),
\end{aligned}
\end{equation}
where $\mathcal{L}_\text{val}^{i}$ and $\mathcal{L}_\text{train}^{i}$ 
denote the training and the validation MLM loss for the i-th language.
Since directly solving this problem can be prohibitive 
due to the expensive inner optimization, 
we approximate $\boldsymbol{\theta}^*$ 
by adapting the current $\boldsymbol{\theta}^{(t)}$ using a single gradient step, 
similar to techniques used in prior meta-learning methods \cite{finn2017model}. 
This results in a two-phase iterative training process 
shown in Algorithm \ref{algorithm} (See Appendix \ref{sec:method_details}).

To be specific, at each training step $t$ on the i-th language during pretraining,
we first adapt a gradient step on $\boldsymbol{\theta}$ 
to obtain a new $\boldsymbol{\theta}'$ 
and update $\phi_i$ based on the $\boldsymbol{\theta}'$'s validation MLM loss:
\begin{equation}
\begin{aligned}
\phi_{i}^{(t+1)} & = \phi_{i}^{(t)} - \alpha \nabla_{\phi_{i}^{(t)}} \frac{1}{L} \sum_{j=1}^{L} \mathcal{L}_\text{val}^{j}(\boldsymbol{\theta}', \phi_{j}^{(t)}) \\
\boldsymbol{\theta}' & = \boldsymbol{\theta}^{(t)} - \beta \nabla_{\boldsymbol{\theta}^{(t)}} \mathcal{L}_\text{train}^{i}(\boldsymbol{\theta}^{(t)}, \phi_{i}^{(t)}),
\end{aligned}
\label{eq:phi}
\end{equation}
where $\alpha$ and $\beta$ are learning rates.
Notice that $\boldsymbol{\theta}'$ is a function of $\phi_{i}^{(t)}$ 
and thus this optimization requires computing the gradient of gradient.
Particularly, by applying chain rule to the gradient of $\phi_{i}^{(t)}$, 
we can observe that it contains a higher-order term:
\begin{small}
\begin{equation}
\left[ \nabla_{\phi_{i}^{(t)},\boldsymbol{\theta}^{(t)}}^2 \mathcal{L}_\text{train}^{i}(\boldsymbol{\theta}^{(t)}, \phi_{i}^{(t)}) \right]  \cdot \left[ \nabla_{\boldsymbol{\theta}'} \frac{1}{L} \sum_{j=1}^{L} \mathcal{L}_\text{val}^{j} (\boldsymbol{\theta}', \phi_{j}^{(t)}) \right]
\label{eq:higher_order_gradient}
\end{equation}
\end{small}
This is important, since it shows that $\phi_{i}$ 
can obtain information from other languages through higher-order gradients.
In other words, language-specific parameters 
are \textbf{not} agnostic of other languages anymore 
without violating the language-specific requirement. 
This is because, in Eq. \ref{eq:phi}, while $\nabla_{\boldsymbol{\theta}^{(t)}}$ 
is based on the $i$-th language only, 
the validation loss is computed for all languages.
Finally, in the second phase, we update $\boldsymbol{\theta}$ based on the new $\boldsymbol{\phi}^{(t+1)}$:
\begin{equation}
\boldsymbol{\theta}^{(t+1)} = \boldsymbol{\theta}^{(t)} - \beta \nabla_{\boldsymbol{\theta}^{(t)}} \mathcal{L}_\text{train}(\boldsymbol{\theta}^{(t)}, \boldsymbol{\phi}^{(t+1)})
\end{equation}

\subsection{Evaluation}
While our method is generic, we evaluate it applied 
on bilingual models with adapter networks.
Adapters have been effectively utilized in multilingual models \cite{bapna2019simple}, 
and we choose them for practical consideration of limiting per-language capacity. 
Unlike prior works that finetune adapters for adaptation, 
we train them jointly with shared parameters during pretraining.
We follow \citet{houlsby2019parameter} and insert language-specific adapters 
after attention and feedforward layers.
We leave a more thorough investigation of 
how to better pick language-specific structures for future work.
For downstream task evaluation, 
we finetune all layers.
Notice that computing the gradient of gradient 
in Eq. \ref{eq:phi} doubles the memory requirement.
In practice, we utilize the finite difference approximation 
(Appendix \ref{sec:method_details}).

By evaluating their performance on the zero-shot transfer settings 
(Table \ref{tab:ner_pos}, \ref{tab:qa} and \ref{tab:xnli}), 
we observe that our method, denoted as meta adpt, 
consistently improves the performance over JointPair baselines,
while ordinary adapters (adpt) perform worse than JointPair.
This shows that, the proposed method can effectively utilize 
the added language-specific adapters to improve generalization 
of shared parameters across languages.
At the same time, our method also mitigates negative interference 
and outperforms JointPair on within-language performance, 
closing the gap with monolingual models.
In particular, it performs better than ordinary adapters in both settings.
We hypothesize that this is because it alleviates language conflicts 
during training and thus converges more robustly.
For example, we plot training loss in the early stage in Figure \ref{fig:train_loss}, 
which shows that ordinary adapters converge slower than JointPair 
due to overfitting of language-specific adapters
while meta adapters converge much faster.
For ablation studies, we also report results for JointPair 
trained with adapters shared between two languages, denoted as share adpt.
Unlike language-specific adapters that can hinder transferability, shared adapters improve both within-language and cross-lingual performance with the extra capacity.
However, meta adapters still obtain better performance.
These results show that mitigating negative interference can improve multilingual representations.

\begin{table}[t!]
\begin{center}
\begin{small}
\resizebox{0.9\columnwidth}{!}{%
\begin{tabular}{l | c c c c c c }
\toprule
Model & ar & fr & ru & hi & sw & avg \\
\toprule
JointPair & 67.1 & 73.5 & 69.2 & 61.5 & 62.3 & 66.7 \\
\enspace + ffn & 42.5 & 51.4 & 40.7 & 36.2 & 34.8 & 41.1\\
\enspace + attn & 48.5 & 50.7 & 41.2 & 33.3 & 35.1 & 41.8 \\
\enspace + adpt & 67.8 & 73.7 & 69.5 & 62.2 & 59.7 & 66.6 \\
\enspace + share adpt & 67.9 & 73.4 & 70.0 & 61.8 & 62.2 & 67.1 \\
\enspace + meta adpt & 68.5 & 74.8 & 70.2 & 64.5 & 61.5 & 67.9 \\
\toprule
XLM & 68.2 & 75.2 & 72.3 & 65.4 & 58.1 & 67.8 \\
\bottomrule
\end{tabular}
}%
\end{small}
\end{center}
\caption[caption]{XNLI results (Accuracy).}
\vskip -0.1in
\label{tab:xnli}
\end{table}

\section{Related Work}

Unsupervised multilingual language models such as mBERT \cite{devlin2018bert} and XLM \cite{lample2019cross,conneau2019unsupervised} work surprisingly well
on many NLP tasks without parallel training signals \cite{pires2019multilingual,wu2019beto}.
A line of follow-up work \cite{wu2019emerging,artetxe2019cross,karthikeyan2020cross} study 
what contributes to the cross-lingual ability of these models.
They show that vocabulary overlap is not required for multilingual models, 
and suggest that abstractions shared across languages emerge automatically during pretraining.
Another line of research investigate how to further improve these shared knowledge, 
such as applying post-hoc alignment \cite{wang2020cross,cao2020multilingual} 
and utilizing better calibrated training signal \cite{mulcaire2019polyglot,huang2019unicoder}.
While prior work emphasize how to share to improve transferability, 
we study multilingual models from a different perspective of 
how to unshare to resolve language conflicts.

Our work is also related to transfer learning \citep{pan2010survey} and multi-task learning \citep{ruder2017overview}.
In particular, prior work have observed \citep{rosenstein2005transfer} and studied \citep{wang2019characterizing} negative transfer, such that transferring knowledge from source tasks can degrade the performance in the target task.
Others show it is important to remedy negative transfer in multi-source settings \citep{ge2014handling,wang2018towards}.
In this work, we study negative transfer in multilingual models, where languages contain heavily unbalanced training data and exhibit complex inter-task relatedness.

In addition, our work is related to methods that measure 
similarity between cross-lingual representations.
For example, existing methods utilize statistical metrics 
to examine cross-lingual embeddings such as singular vector 
canonical correlation analysis \cite{raghu2017svcca,kudugunta2019investigating}, 
eigenvector similarity \cite{sogaard2018limitations}, 
and centered kernel alignment \cite{kornblith2019similarity,wu2019emerging}.
While these methods focus on testing latent representations,
we directly compare similarity of neural network structures through network pruning.
Finally, our work is related to meta learning, 
which sets a meta task to learn model initialization for fast adaptation \cite{finn2017model,gu2018meta,flennerhag2019meta}, 
data selection \cite{wang2020balancing}, 
and hyperparameters \cite{baydin2018online}.
In our case, the meta task is to mitigate negative interference.

\section{Conclusion}

We present the first systematic study of negative interference 
in multilingual models and shed light on its causes.
We further propose a method and show it can improve cross-lingual transferability 
by mitigating negative interference. 
While prior efforts focus on improving sharing and cross-lingual alignment, 
we provide new insights and a different perspective 
on unsharing and resolving language conflicts.

\section*{Acknowledgments}
We want to thank Jaime Carbonell for his support in the early stage of this project.
We also would like to thank Zihang Dai, Graham Neubig, Orhan Firat, Yuan Cao, Jiateng Xie, Xinyi Wang, Ruochen Xu and Yiheng Zhou for insightful discussions. 
Lastly, we thank anonymous reviewers for their valuable feedback.

\bibliography{emnlp2020}
\bibliographystyle{acl_natbib}

\clearpage

\appendix

\begin{table*}[t!]
\begin{center}
\begin{tabular}{l | c c c c c }
\toprule
    Model & ar  & ru  & sw & te & avg \\
\toprule
\multicolumn{6}{c}{Within-language Monolingual}\\
\toprule
Mono & 74.2/62.5 & 63.1/49.2 & 52.5/37.4 & 58.2/41.0 & 62.0/47.5 \\
JointPair & 71.3/58.1 & 58.2/43.1 & 52.8/39.0 & 52.2/36.4 & 58.6/44.2  \\
\enspace + ffn & 73.4/61.2 & 61.2/45.8 & 51.4/34.3 & 57.5/40.5 & 60.9/45.5 \\
\enspace + attn & 72.8/61.0 & 60.8/45.4 & 51.2/34.0 & 52.8/36.8 & 59.4/44.3 \\
\enspace + adpt & 71.5/58.7 & 59.4/44.8 & 52.1/38.7 & 55.5/38.9 & 59.6/45.3 \\
\enspace + share adpt & 71.0/57.8 & 58.5/43.2 & 52.8/39.0 & 53.9/37.2 & 59.1/44.3 \\ 
\enspace + meta adpt & 73.0/61.4 & 61.8/46.7 & 54.5/40.0 & 56.2/39.5 & 61.4/36.4 \\
\toprule
XLM & 74.3/63.2 & 62.5/48.7 & 58.7/40.4 & 55.4/38.3 & 62.7/47.7 \\
\toprule
\multicolumn{6}{c}{Zero-shot Cross-lingual}\\
\toprule
JointPair & 54.1/39.5 & 43.2/27.5 & 41.5/22.2 & 21.5/14.7 & 40.1/26.0 \\
\enspace + ffn & 2.2/1.5 & 0.0/0.0 & 4.4/3.7 & 0.0/0.0 & 1.7/1.3 \\
\enspace + attn & 3.7/2.0 & 2.1/1.2 & 0.7/1.0 & 0.0/0.0 & 1.6/1.1 \\
\enspace + adpt & 53.4/39.1 & 44.7/27.9 & 41.2/21.8 & 20.4/13.8 & 39.9/25.7 \\
\enspace + share adpt & 54.3/39.6 & 44.8/27.8 & 42.2/22.9 & 22.7/15.6 & 41.0/26.5 \\
\enspace + meta adpt & 57.5/40.8 & 45.8/28.8 & 43.0/24.2 & 23.1/17.7 & 42.4/27.9 \\
\toprule
XLM & 59.4/41.2 & 47.3/29.8 & 42.3/22.0 & 16.3/7.2 & 41.3/25.1 \\
\bottomrule
\end{tabular}%
\end{center}
\caption[caption]{Full results on TyDiQA-GoldP (F1/EM).}
\vskip -0.1in
\label{tab:qa_full}
\end{table*}

\section{Fine-tuning Details}
\label{sec:finetune_details}

Notice that XNLI only has training data in available in English so we only evaluate zero-shot cross-lingual performance on it.
Following \cite{hu2020xtreme}, we finetune the model for 10 epochs for NER and POS, 2 epochs for QA and 200 epochs for XNLI.
For NER, POS and QA, we search the following hyperparameters: batch size \{16, 32\}; learning rate \{2e-5, 3e-5, 5e-5\}.
We use English dev set for zero-shot cross-lingual setting and the target language dev set for within-language monolingual setting.
For XNLI, we search for: batch size \{4, 8\}; encoder learning rate \{1e-6, 5e-6, 2e-5\}; classifier learning rate \{5e-6, 2e-5, 5e-5\}.
For models with language-specific components, we test freezing these components or finetuning them together.
We discover that finetuning the whole network always yields better results.
For all experiments, we save checkpoint after each epoch.

\section{Method Details}
\label{sec:method_details}

Let $z_i$ be the output of the $i$-th layer of dimension $d$. The residual adapter network \cite{rebuffi2017learning,houlsby2019parameter,bapna2019simple} is a bottleneck layer that first projects $z_i$ to an inner layer with dimension $b$:
\begin{equation}
    h_i = g(W^z_i z_i)
\end{equation}
where $W^z_i \in \mathbb{R}^{d \times b}$ and $g$ is some activation function such as $relu$. It is then projected back to the original input dimension $d$ with a residual connection:
\begin{equation}
    o_i = W^h_i h_i + z_i
\end{equation}
where $W^h_i \in \mathbb{R}^{b \times d}$. In our experiments, we fix $b = \frac{1}{4} d$. For a bilingual model of $\text{\it lg}_1$ and $\text{lg}_2$, we inject two langauge-specific adapters after each attention and feedforward layer, one for each language. For example, if the input text is in $\text{lg}_1$, the network will be routed to adapters designated for $\text{lg}_1$. The rest of the network and training protocol remain unchanged.

The injected adapter layers mimic the warp layers interleaved between base network layers in \citet{flennerhag2019meta}. Warp layers are meta parameters that aim to improve the performance of the base network. They precondition base network gradients to obtain better gradient geometry. In our experiments, we treat language-specific adapters as meta parameters to improve generalization of the shared network.
The algorithm is outlined in Algorithm \ref{algorithm}.
The adapters are updated according to Eq \ref{eq:phi}, which doubles the memory requirement. In particular, the high-order term in Eq \ref{eq:higher_order_gradient} requires computing the gradient of gradient. In practice, we approximate this term using the finite difference approximation as:
\begin{equation}
\frac{\nabla_{\phi_{i}^{(t)}} \mathcal{L}_\text{train}^{i}(\boldsymbol{\theta}^{+}, \phi_{i}^{(t)}) - \nabla_{\phi_{i}^{(t)}}  \mathcal{L}_\text{train}^{i}(\boldsymbol{\theta}^{-}, \phi_{i}^{(t)})}{2 \epsilon}
\end{equation}
where $\boldsymbol{\theta}^{\pm} = \boldsymbol{\theta}^{(t)} \pm \epsilon \nabla_{\boldsymbol{\theta}'} \frac{1}{L} \sum_{j=1}^{L} \mathcal{L}_\text{val}^{j} (\boldsymbol{\theta}', \phi_{j}^{(t)})$ and $\epsilon$ is a small scalar.
We use the same value for learning rates $\alpha$ and $\beta$ in Eq \ref{eq:phi}, to be consistent with standard learning rate schedule used in XLM \cite{lample2019cross}.

\section{Extra Results}
\label{sec:extra_results}

We show the full results on the TyDiQA-GoldP dataset in Table \ref{tab:qa_full}.

\end{document}